\documentclass[10pt,twocolumn,letterpaper]{article}

\usepackage[pagenumbers]{cvpr} %
\usepackage[dvipsnames]{xcolor}

\definecolor{cvprblue}{rgb}{0.21,0.49,0.74}
\definecolor{cPLOT_gc_ppsm}{RGB}{127, 201, 127}
\definecolor{cPLOT_sm_comb}{RGB}{253, 192, 10}
\definecolor{cPLOT_dpfm}{RGB}{240, 2, 127}
\definecolor{cPLOT_ours}{RGB}{20, 120, 120}

\usepackage{overpic}
\usepackage[dvipsnames, table]{xcolor}
\usepackage{times}
\usepackage{epsfig}
\usepackage{graphicx}
\usepackage{amsmath}
\usepackage{amssymb}
\usepackage{amsthm}
\usepackage{mathtools}
\usepackage{bm}
\makeatletter
\@namedef{ver@everyshi.sty}{}
\makeatother
\usepackage[mode=buildnew]{standalone}
\usepackage{tikz}
\usetikzlibrary{positioning}
\usetikzlibrary{shapes} 
\usetikzlibrary{matrix}
\usetikzlibrary{calc} 
\usetikzlibrary{intersections}
\usepackage[accsupp]{axessibility}  %
\usepackage{tikzscale}

\usepackage{pgfplots}
\usepackage[eulergreek]{sansmath}
\usepackage{caption,subcaption}
\usepackage{bbm}
\usepackage{comment}
\usepackage{acro}
\usepackage{listings}
\usepackage{multirow}
\usepackage{booktabs}
\usepackage{wrapfig}
\usepackage{algorithm}
\usepackage{algpseudocode}
\usepackage{float}
\usepackage{overpic}
\usepackage{cuted}
\usepackage{pbox}
\usepackage{tabularx}
\setkeys{Gin}{keepaspectratio}
\usepackage{adjustbox}
\usepackage{enumitem}

\newcommand{\verts}{\boldsymbol{V}}
\newcommand{\triangles}{\boldsymbol{F}}
\newcommand{\numTriangles}[1]{{|\boldsymbol{F}_{#1}|}}
\newcommand{\edges}{\boldsymbol{E}}
\newcommand{\shapeX}{{\boldsymbol{X}}}
\newcommand{\shapeY}{{\boldsymbol{Y}}}
\newcommand{\contour}{\boldsymbol{C}}
\newcommand{\prodGraph}{{\boldsymbol{P}}}

\newcommand{\twovector}[2]{\bigl(\begin{smallmatrix}#1 \\#2\end{smallmatrix}\bigr)}

\newcommand{\boundary}{\text{bn}}
\newcommand{\interior}{\text{in}}

\newcommand{\injectivitySlack}{s^\text{inj}}
\newcommand{\surjectivitySlack}{s^\text{sur}}
\newcommand{\shortNewMetric}{GeoED}

\newcommand{\matchingCost}{{c}}
\newcommand{\overlapX}{o_\shapeX}
\newcommand{\overlapY}{o_\shapeY}

\def\lowres{\text{lr}}
\def\AllowedMatches{\boldsymbol{A}}
\def\highToLowresX{\gamma_\shapeX}
\def\highToLowresY{\gamma_\shapeY}

\definecolor{cvprblue}{rgb}{0.21,0.49,0.74}
\usepackage[pagebackref,breaklinks,colorlinks,citecolor=cvprblue]{hyperref}
\usepackage[capitalize]{cleveref}
\newtheorem{theorem}{Theorem}

\newtheorem{definition}[theorem]{Definition}

\def\paperID{158} %
\def\confName{3DV\xspace}
\def\confYear{2026\xspace}
\newcommand{\authorspace}{\hspace{0.7cm}}
\newcommand{\affiliationspace}{\hspace{0.6cm}}
\title{An Integer Linear Programming Approach to \\
Geometrically Consistent Partial-Partial Shape Matching}

\author{
Viktoria Ehm$^{1,2}$
\authorspace
Paul Roetzer$^{3,4}$ 
\authorspace
Florian Bernard$^{3,4}$
\authorspace
Daniel Cremers$^{1,2}$ \\
$^1$ Technical University of Munich 
\affiliationspace 
$^2$ MCML
\affiliationspace
$^3$ University of Bonn
\affiliationspace
$^4$ Lamarr Institute
}

\begin{document}
\maketitle

\newlength{\teaserheight}
\setlength{\teaserheight}{3.3cm}
\newlength{\runtimeheight}
\setlength{\runtimeheight}{1.0cm}
\begin{strip}
    \centering
    \footnotesize%
    \begin{tabular}{cc}
        \includegraphics[height=\teaserheight, trim={0 0 0 0.7cm}]{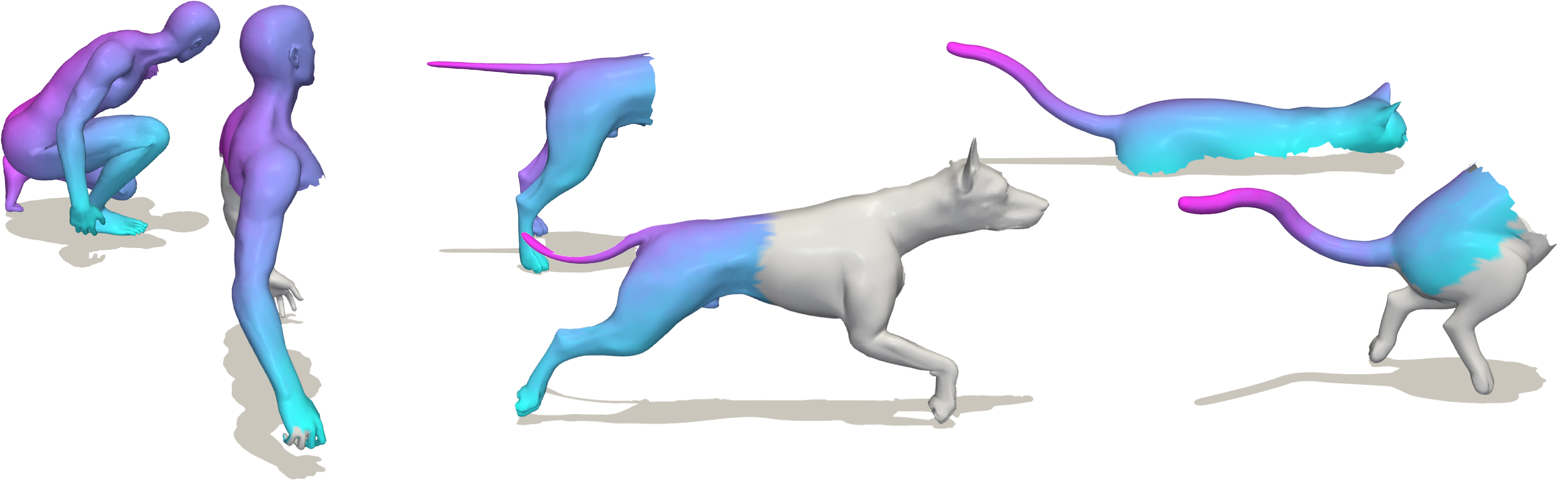}
        &
        \hspace{-0.6cm}
        \begin{tabular}{c}
            \adjustbox{trim=0 0.2cm 0 \teaserheight}{%
                 \newcommand{\rtLineWidth}{1.5pt}
\newcommand{\plotWidth}{0.48\linewidth}
\newcommand{\plotHeight}{0.3\linewidth}
\newcommand{\rtTitle}{\textsc{Runtime of Geo.~Consistent Methods}}

\pgfplotsset{%
    label style = {font=\small},
    tick label style = {font=\small},
    title style =  {font=\normalsize},
    legend style={  fill= gray!10,
                    fill opacity=0.6, 
                    font=\small,
                    draw=gray!20, %
                    text opacity=1}
}
\begin{tikzpicture}[scale=0.7, transform shape]
	\begin{axis}[
		width=\plotWidth,
		height=\plotHeight,
		grid=major,
		title=\rtTitle,
		legend style={
			at={(0.98,0.98)},
			anchor=north east,
			legend columns=1},
		legend cell align={left},
	ylabel=\small{Runtime in Minutes},
        xmin=0,
        xmax=700,
        xlabel=\small{Total Number of Triangles},
        ylabel near ticks,
        xtick={0,100, 200, 300, 400, 500, 600, 700},
        ymin=0,
        ymax=120,
        ytick={0, 30, 60, 90, 120},
	]
\addplot [color=cPLOT_gc_ppsm, solid, smooth, line width=\rtLineWidth]
table[row sep=crcr]{%
0.0  0.0\\
75   97.79\\
100  203.66\\
    };
\addlegendentry{\textcolor{black}{GCPPSM~\cite{ehm2024partial}}}
\addplot [color=cPLOT_ours, solid, smooth, line width=\rtLineWidth]
table[row sep=crcr]{%
0 0 \\
100 0.26 \\
200 0.95 \\
300 4.07 \\
400 1.46 \\
500 3.24 \\
600 4.13 \\
700 10.28 \\
800 12.55 \\
900 16.76 \\
1000 15.02 \\
    };
\addlegendentry{\textcolor{black}{Ours}}
\end{axis}
\end{tikzpicture}

            }
        \end{tabular}
        \\
   \end{tabular}
    \vspace{-0.1cm}
    
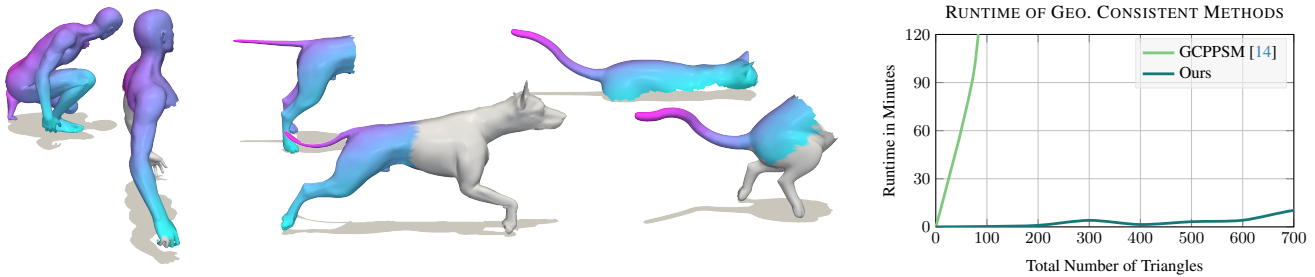
\captionof{figure}{
        \textbf{(Left)} We show matching results for pairs of partial shapes computed with our method using colour transfer.
        \textbf{(Right)} Runtime comparison to the recent (geometrically consistent) partial-partial shape matching method GC-PPSM~\cite{ehm2024partial} which approaches the task using integer non-linear programming.
        In contrast, we propose the first integer \emph{linear} programming approach explicitly tailored to partial-partial shape matching with geometric consistency, making our method more scalable in comparison.
    }
    \label{fig:teaser}
\end{strip}

\begin{abstract}
The task of establishing correspondences between two 3D shapes is a long-standing challenge in computer vision. 
While numerous studies address full-full and partial-full 3D shape matching, only a limited number of works have explored the partial-partial setting, very likely due to its unique challenges: we must compute accurate correspondences while at the same time find the unknown overlapping region.
Nevertheless, partial-partial 3D shape matching reflects the most realistic setting, as in many real-world cases, such as 3D scanning, shapes are only partially observable. 
In this work, we introduce the first integer linear programming approach specifically designed to address the distinctive challenges of partial-partial shape matching. 
Our method leverages geometric consistency as a strong prior, enabling both robust estimation of the overlapping region and computation of neighbourhood-preserving correspondences. 
We empirically demonstrate that our approach achieves high-quality matching results both in terms of matching error and smoothness. 
Moreover, we show that our method is more scalable than previous formalisms.
Our code is publicly available at \url{https://github.com/vikiehm/partial-geco}.

\end{abstract}
    
\section{Introduction}
\label{sec:intro}
Solving correspondence problems is a fundamental prerequisite for a wide range of tasks in computer vision, as many downstream applications depend on the ability to relate data points or features across different observations.  
Prominent examples include 3D reconstruction~\cite{leroy2024grounding}, where accurate correspondences enable the fusion of multiple views into a coherent geometric model, object recognition~\cite{kim19913}, where matching enables identification of known shapes, tracking~\cite{schoenemann2009combinatorial}, where correspondences across time are essential, and image matching~\cite{ma2021image}, which underlies applications such as registration and stitching.  
Beyond that, correspondence problems arise in shape analysis,  such as in 2D shape matching~\cite{schmidt2009planar}, in 2D-3D shape matching~\cite{lahner2016efficient, roetzer2023conjugate}, and in 3D shape matching~\cite{windheuser2011geometrically, rodola2012game, solomon2012soft, ovsjanikov2012functional, ezuz2019elastic, bernard2020mina, cao2023unsupervised}. 
In real-world scenarios, however, the acquisition of 3D shapes often relies on scanning devices that are limited in their ability to capture a complete surface, resulting in shapes that are only partially observed \cite{cui2012algorithms}.  
Such partial observations can occur due to occlusions, limited sensor range, or constraints on scanning time, making complete coverage impractical or impossible.  
Consequently, the most realistic and practically relevant setting for 3D shape matching is the partial-partial case, in which both shapes are incomplete.  
Despite its relevance, this scenario remains underexplored, with only a small number of works explicitly addressing it.  
The main difficulty lies in its combined challenges: we must not only establish reliable correspondences between the observed regions, but also infer the a priori unknown spatial overlap between the shapes.
These two requirements greatly increase the problem's complexity compared to full-full or partial-full settings, where the overlap for at least one shape is known.  
To address these challenges effectively, powerful regularisation strategies are necessary.
As such, geometric consistency, in the form of neighbourhood preservation, serves as a particularly strong prior for regularising the resulting correspondences.  
Nevertheless, geometric consistency for partial-partial 3D shape matching has so far only been implemented within a non-linear programming framework~\cite{ehm2024partial}, which can limit scalability and computational efficiency, cf.~\cref{fig:teaser} right.  
In this work, we build upon the recently proposed shape matching formalism~\cite{roetzer2025geco} and introduce the first geometrically consistent integer \emph{linear} programming formalism explicitly tailored for partial-partial 3D shape matching.  
Our formulation is designed to exploit geometric consistency not only to identify plausible overlap regions but also to compute correspondences that preserve local neighbourhood relations.
In turn, this leads to empirically observed high-quality matching results as well as robust overlap predictions.
In particular, we summarise our main contributions as follows:
\begin{itemize}
    \item We introduce the first integer \emph{linear} programming formalism for partial-partial shape matching that explicitly incorporates geometric consistency.  
    \item For the first time, our framework enables the integration of a predicted overlap region into the matching process.  
    \item Through experiments, we demonstrate that our formalism scales significantly better than previous geometrically consistent partial-partial approaches. %
    \item We further show that our method achieves high-quality matching results in terms of geodesic errors, smoothness, and overlap scores. 
\end{itemize}

\section{Related Works}
\label{sec:related-works}
In this section, we review works which we consider most relevant to our paper.
For a broader overview of shape matching methods, we invite the interested reader to read the survey papers~\cite{van2011survey,tam2012registration,sahilliouglu2020recent}.

\textbf{Partial Shape Matching.}
Matching of complete shapes, i.e.~full-full shape matching has received lots of attention in the past years.
In contrast, partial shape matching, i.e.~shape matching where at least one shape is incomplete, is less explored (partial-full shape matching) or is even underexplored (partial-partial shape matching) due to yet-to-overcome challenges~\cite{ehm2025beyond}.
In one of the first works of partial-full shape matching, the problem is tackled by using functional maps~\cite{ovsjanikov2012functional} and by exploiting the resulting slanted diagonal structure of respective functional maps~\cite{rodola2017partial}.
This idea was further used as a loss function in supervised~\cite{attaiki2021dpfm} and unsupervised~\cite{attaiki2021dpfm, cao2022unsupervised, cao2023unsupervised} learning frameworks.
Other lines of work focus on preserving distances between points to act as regularisation when points are being matched from source to target shape~\cite{bracha2023partial, bracha2024wormhole} or incorporate neighbourhood preservation as a hard constraint~\cite{ehm2024geometrically, roetzer2025geco}.
Among the few works specifically tailored to the partial-partial problem, predicting the overlapping region in learning-based methods is approached using cross-attention~\cite{attaiki2021dpfm} or so-called correspondence reflections~\cite{xie2025echomatch}, which both yield probabilities stating if a vertex is or is not within the overlapping region.
In contrast, axiomatic methods rely on non-linear integer programming and neighbourhood preservation constraints to determine the overlapping region~\cite{ehm2024partial}.
Our formalism explicitly allows to incorporate such overlapping region probabilities into the cost function, while at the same time it regularises the overlapping region by enforcing neighbourhood preservation.

\textbf{Geometrically Consistent Shape Matching.}
Geometric consistency, i.e.~the preservation of neighbourhood relations of surface elements, can act as a strong regulariser to compute high-quality shape matching results and can resolve ambiguities in ambiguous matching problems such as partial-partial shape matching. 
Yet, geometric consistency is often neglected due to resulting hard-to-solve formalisms.
As such, some approaches incorporate neighbourhood information using formalisms based on the quadratic assignment problem, which, due to its NP-hardness~\cite{rendl1994quadratic}, can only be solved approximately~\cite{burghard2017efficient, kushinsky2019sinkhorn, holzschuh2020simulated, solomon2016entropic}.
Other lines of work require an initial set of (sparse) correspondences and either deform one shape to the other~\cite{amberg2007optimal} or match shapes via intermediate domains~\cite{baden2018mobius, schmidt2019distortion, schmidt2020inter}.
In contrast, Windheuser et al.~\cite{windheuser2011geometrically}, have proposed an initialisation-free integer linear program based upon matching triangles and preserving their neighbourhood.
Due to the complexity of the resulting optimisation problem, it can only be solved using coarse-to-fine strategies~\cite{windheuser2011large} or approximate solvers~\cite{roetzer2022scalable, roetzer2024discomatch}.
Later, in SpiderMatch~\cite{roetzer2024spidermatch}, authors propose to represent the source shape using a self-intersecting, cyclic curve and match this curve to the target shape while preserving the intersections of the curve. 
While this formalism is more scalable than previous works, it only enforces geometric consistency at intersection points. 
Building on the idea of alternative 3D shape representations, the recent work GeCo~\cite{roetzer2025geco} represents the source shape using multiple cyclic curves and matches these curves to the target shape while preserving the neighbourhood of the individual cycles.
This results in better geometric consistency compared to SpiderMatch~\cite{roetzer2024spidermatch} while remaining similarly scalable.
A follow-up work~\cite{amrani_high}, building on GeCo~\cite{roetzer2025geco}, showed that the formalism can be adapted to match shapes with up to 10k triangles.
We also build our work on GeCo~\cite{roetzer2025geco} and propose a novel formalism which is explicitly designed for partial-partial shape matching.

\textbf{Geometrically Consistent Partial-Partial Shape Matching.}
Partial-partial shape matching remains underexplored and usually geometric consistency is neglected due to hard-to-solve formalisms, even though it can act as a strong regulariser to matching problems. 
Consequently, there exists only one work which tackles partial-partial shape matching with geometric consistency: 
\cite{ehm2024partial}~proposes a non-linear integer programming approach which builds on the formalism introduced in~\cite{windheuser2011geometrically} for geometrically consistent full-full shape matching.
As such, \cite{ehm2024partial} leverages geometric consistency to compute the overlapping region.
Yet, due to non-linearity, the resulting formalism does not scale well, see \cref{fig:teaser} right.

In our work, we propose an integer \emph{linear} programming formalism which 
also uses geometric consistency as a prior to compute the overlapping region.
In addition, since our formalism minimises a linear objective compared to a non-linear one~\cite{ehm2024partial}, our approach is more scalable.

\section{Background}
\label{sec:background}
In the following, we discuss relevant background for our method: we define shapes and an alternative shape representation~(\cref{sec:background-shapes}), introduce product graphs~(\cref{sec:background-productgraphs}) and define geometric consistency for partial-partial shape matching~(\cref{sec:background-p2p-geo-cons}).
Our notation is summarised in \cref{table:notation}.

\begin{table}
    \renewcommand{\arraystretch}{0.9}
    \setlength{\tabcolsep}{0.25em}
        \small\centering
	\begin{tabularx}{\columnwidth}{lp{5.6cm}}
        \toprule
        \textbf{Symbol} & \textbf{Description} \\
        \toprule
        $\shapeX = (\verts_\shapeX,\edges_\shapeX,\triangles_\shapeX)\!\!\!$ &3D shape\\
        $\verts_\shapeX = \verts_\shapeX^\boundary \cup \verts_\shapeX^\interior$ & Vertices of shape $\shapeX$ with boundary $\verts_\shapeX^\boundary$\\
        & and inner $\verts_\shapeX^\interior$ vertices $(\verts_\shapeX^\boundary\cap\verts_\shapeX^\interior=\emptyset)$\\       
        $\edges_\shapeX$ &Edges of shape $\shapeX$\\
        $\triangles_\shapeX$ &Triangles of shape $\shapeX$\\
        $\shapeY$ &3D shape $\shapeY$ (defined analogously to $\shapeX$)\\
        $\contour_i$ & $i$-th surface cycle (representing the $i$-th\\
                     &triangle $f_i\in\triangles_\shapeX$ if shape $\shapeX$)\\
        $\prodGraph_i = (\verts_{\prodGraph_i}, \edges_{\prodGraph_i})$ & Product graph between $i$-th \\
                    & surface cycle $\contour_i$ and 3D shape $\shapeY$\\
        $\prodGraph= (\verts_{\prodGraph}, \edges_{\prodGraph})$ & Collection of all $i=1\dots\numTriangles{\shapeX}$\\
                    & product graphs $\prodGraph_i$\\
        $\matchingCost_k$ &Matching cost for product edge $e_k\in\edges_\prodGraph$\\
        $\overlapX$ & Overlap probability of edges of shape $\shapeX$\\
        $\overlapY$ & Overlap probability of vertices of shape $\shapeY$\\
        \bottomrule
	\end{tabularx}
        \vspace{-0.2cm}
	\caption{Summary of the \textbf{notation} used in this paper.
	}
	\label{table:notation}
\end{table}

\subsection{3D Shapes and Shape Representations}\label{sec:background-shapes}

For our formalism, we consider partial 3D shapes (i.e.~oriented 2D manifold in 3D space with boundary) which we define as follows:

\begin{definition}[3D Shape]
    A 3D shape $\shapeX=(\verts_\shapeX, \edges_\shapeX, \triangles_\shapeX)$ is a triplet consisting of vertices $\verts_\shapeX$, directed edges $\edges_\shapeX\subset\verts_\shapeX\times\verts_\shapeX$ and oriented triangles $\triangles_\shapeX\subset\verts_\shapeX\times\verts_\shapeX\times\verts_\shapeX$ such that $\shapeX$ forms an oriented 2D manifold in 3D space with potential boundaries. 
    Consequently, the set of vertices $\verts_\shapeX = \verts_\shapeX^\boundary \cup \verts_\shapeX^\interior$ can be partitioned into boundary vertices $\verts_\shapeX^\boundary$ and non-boundary vertices $\verts_\shapeX^\interior$ such that $\verts_\shapeX^\boundary\cap\verts_\shapeX^\interior=\emptyset$.
 \end{definition}

Directed edges mean that edges have an orientation, i.e.~whenever an edge $e$ is in the set of edges $e\in\edges$ does not imply that its opposite edge $-e$ is in the set of edges, i.e.~$e\notin\edges$.

 Following \cite{roetzer2025geco}, we consider an alternative 3D shape representation using what they call \emph{surface cycles}.
\begin{definition}[Surface Cycle Shape Representation]\label{def:surface-cycles}
    The triangles $\triangles_\shapeX$ of a 3D shape $\shapeX$ can be represented with a collection of $\numTriangles{\shapeX}$-many surface cycles $\contour_1,\dots,\contour_\numTriangles{\shapeX}$ such that the $i$-th triangle $f_i\in\triangles_\shapeX$ is represented with the $i$-th surface cycle $\contour_i$. The $i$-th surface cycle $\contour_i=(\verts_{\contour_i},\edges_{\contour_i})$ is a cyclic chain graph consisting of the three vertices $v_1,v_2,v_3\in\verts_{\contour_i}\subset\verts_\shapeX$ and the three oriented edges $e_1,e_2,e_3\in\edges_{\contour_i}\subset\edges_\shapeX$ of the $i$-th triangle.
\end{definition}
The intuition is that every oriented triangle of shape $\shapeX$ forms a directed cyclic chain graph with three edges and three vertices such that the surface of $\shapeX$ is tiled into $\numTriangles{\shapeX}$-many surface cycles which are glued together at opposite edges, see~\cite{roetzer2025geco} and \cref{fig:overview} for an illustration.

\subsection{Shape Matching with Product Graphs}\label{sec:background-productgraphs}

Shortest paths in product graphs have been introduced for various shape matching tasks, including shape-image matching~\cite{schoenemann2009combinatorial}, 2D-3D shape matching~\cite{lahner2016efficient, roetzer2023conjugate, roetzer2025higherorder} and 3D shape matching~\cite{roetzer2024spidermatch, roetzer2025higherorder, roetzer2025geco}. 
The key idea is to compute the product graph $\prodGraph_i$ (or multiple product graphs~\cite{roetzer2025geco}) between a cyclic chain graph $\contour_i$ and another graph, i.e.~a 3D shape $\shapeY$. 
We follow the definition presented in~\cite{lahner2016efficient} in which a product graph $\prodGraph$ reads:
\begin{definition}[Product graph]\label{def:prodgraph}
    The product graph $\prodGraph = (\verts_{\prodGraph}, \edges_{\prodGraph})$ between a cyclic chain graph $\contour$ and a 3D shape $\shapeY$ is a directed graph defined as
    \begin{equation}\label{eq:prod-graph}
    \begin{aligned}
        \verts_{\prodGraph_i} &= \verts_{\contour_i} \times \verts_\shapeY,\\
        \edges_{\prodGraph_i} &= \{\left( v, \bar{v}\right) \in \verts_{\prodGraph_i}\times \verts_{\prodGraph_i}\; |\; 
        v = \twovector{x}{y},\; \bar{v} = \twovector{\bar{x}}{\bar{y}}\\
        &\qquad(x, \bar{x}) \in \edges_{\contour_i},(y, \bar{y}) \in \edges_\shapeY^+\}.
    \end{aligned}
    \end{equation}
    Here $\edges^+_\shapeY := \edges_\shapeY \cup \{(y, y) \; | \; y \in \verts_\shapeY\}$ are the edges of shape $\shapeY$ extended with self-edges.
\end{definition}
In a nutshell, the product graph $\prodGraph$ contains every possible combination between edges of the cyclic chain graph $\contour$ and edges/vertices of shape $\shapeY$ and thus encodes every potential matching between these elements. 
Furthermore, the product graph $\prodGraph$ is defined via the connectivity of vertices of the cyclic chain graph $\contour$ and the connectivity of vertices of the shape $\shapeY$ such that a cyclic path within $\prodGraph$ yields a cyclic path on $\shapeY$, see~\cite{lahner2016efficient, roetzer2025geco} for illustrations and further discussion.
We can define suitable matching costs for every edge in the product graphs (e.g.~vertex-wise feature differences as done in~\cite{lahner2016efficient, roetzer2023conjugate, roetzer2024spidermatch, roetzer2025higherorder, roetzer2025geco}). 
According to these costs we find a shortest, i.e.~minimum cost, cyclic path through the product graph $\prodGraph$ (by variants of Dijkstra's algorithm).
This shortest cyclic path in $\prodGraph$ in turn yields a globally optimal matching between a cyclic chain graph $\contour$ and a 3D shape $\shapeY$ (since every edge in $\prodGraph$ encodes a potential matching).
Yet, for our purposes of geometrically consistent 3D shape matching we want to define multiple of these product graphs and we want to employ additional constraints.
Hence, we cannot use vanilla shortest path algorithms but rather have to consider the integer linear programming formalism of the above explained shortest path problem.

\subsection{Partial-Partial Geometric Consistency}\label{sec:background-p2p-geo-cons}

We aim to find geometrically-consistent matchings between two partial shapes.
This means that for every point in the resulting overlapping region (i.e.~the region on both shapes which is matched) we want to preserve its neighbourhood.
Formally, we define geometric consistency for partial-partial shape matching as follows

\begin{definition} [Partial-Partial Geometric Consistency]
    We call a mapping $\sigma : \verts_\shapeX \rightarrow \verts_\shapeY$ between two partial shapes $\shapeX$ and $\shapeY$ geometrically consistent if for every interior pair of matched points $x,\bar{x}\in\verts_\shapeX^\interior$, that is connected by an edge, i.e.~$(x,\bar{x})\in\edges_\shapeX$ and $\sigma(x) \neq \emptyset$, $\sigma(\bar{x}) \neq \emptyset$ it holds that either $(\phi(x), \phi(\bar{x})) \in \edges_\shapeY$ or $\phi(x) = \phi(\bar{x})$ whenever $\phi(x),\phi(\bar{x})$ map to interior vertices, i.e.~$\phi(x),\phi(\bar{x})\in \verts_\shapeY^\interior$.
\end{definition}

In a nutshell, every pair of points on shape $\shapeX$, that is connected by an edge, should be matched to another pair of points on shape $\shapeY$ which are either connected by an edge or the same vertex.

\section{Partial-Partial Shape Matching}
\label{sec:method}

\begin{figure}
    \centering
    \includegraphics[width=0.8\columnwidth]{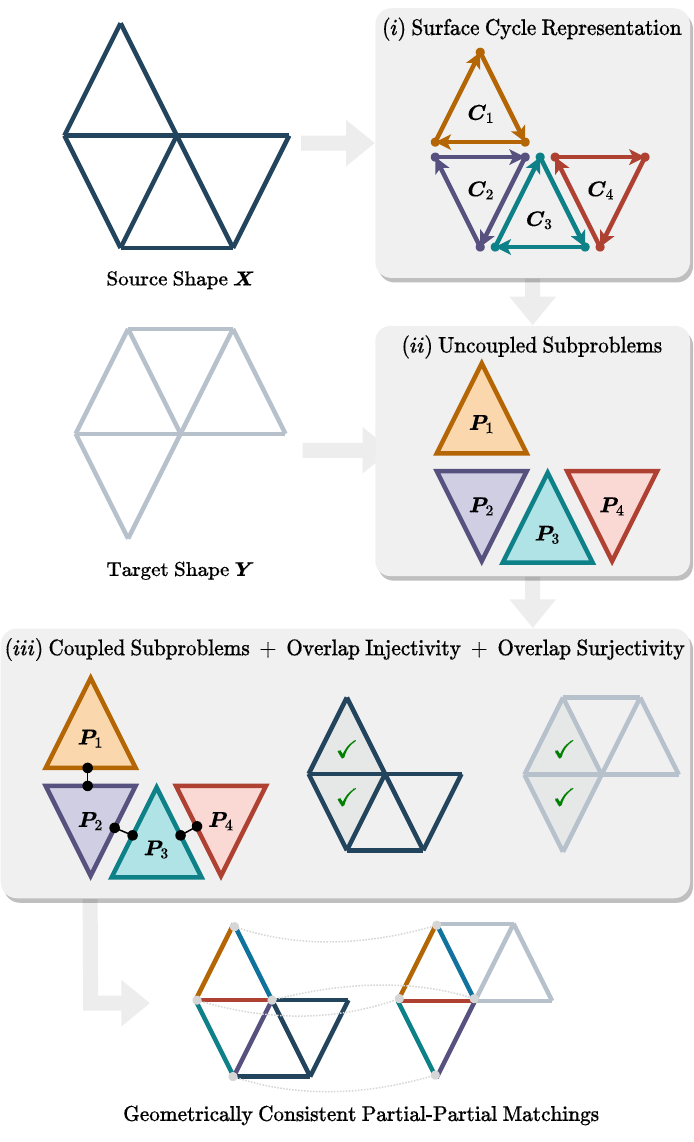}
    \caption{\textbf{Overview} over our partial-partial 3D shape matching approach. We match a partial source 3D shape $\shapeX$ to a non-rigidly deformed partial target 3D shape $\shapeY$. $(i)$ We represent the source shape using surface cycles (cf.~\cref{def:surface-cycles}). $(ii)$ For each surface cycle $\contour_i$, we formulate an independent subproblem as the product graphs $\prodGraph_i$ between a $\contour_i$ and the whole 3D shape $\shapeY$ (cf.~\cref{def:prodgraph}). $(iii)$ By employing coupling constraints (cf.~\cref{sec:coupled-matching-problems}) as well as injectivity and surjectivity constraints of the overlap (cf.~\cref{sec:inj-surj}) we obtain a formalism for geometrically consistent partial-partial 3D shape matching (cf.~\cref{sec:resulting-ilp}).
}
    \label{fig:overview}
\end{figure}

Our goal is to find a geometrically consistent matching between a partial source shape $\shapeX$ and a partial target shape $\shapeY$ (i.e.~two 3D shapes with boundary and unknown overlap).
To this end, we build on~\cite{roetzer2025geco} and consider an independent matching problem for each triangle, which we couple at opposite edges to enforce geometric consistency~(\cref{sec:coupled-matching-problems}).
Furthermore, we design injectivity and surjectivity constraints such that we can account for the unknown overlap between both shapes, i.e.~we allow vertices of both shapes not to be matched~(\cref{sec:inj-surj}).
This leads to our novel integer linear program (ILP) for partial-partial 3D shape matching, which is able to incorporate fuzzy overlap predictions~(\cref{sec:resulting-ilp}).
Finally, we discuss a coarse-to-fine strategy to scale to higher resolutions~(\cref{sec:coarse-to-fine}).
An overview of our approach can be found in \cref{fig:overview}.

\subsection{Coupled Surface Cycle Matching Problems}\label{sec:coupled-matching-problems}

We follow~\cite{roetzer2025geco} and represent each of the $i=1\dots\numTriangles{\shapeX}$ triangles $f_i\in\triangles_\shapeX$ of source shape $\shapeX$ with a surface cycles $\contour_i$, cf.~\cref{def:surface-cycles}.
For each surface cycle $\contour_i$, we compute a product graph $\prodGraph_i$ between $\contour_i$ and the target 3D shape $\shapeY$ which gives us an indepent subproblem allowing us to match the $i$-th triangle of shape $\shapeX$ to shape $\shapeY$. 
We collect all product graphs $\prodGraph_i$ in what we call the \emph{collection of product graphs} $\prodGraph$. 

\begin{definition}[Collection of Product Graphs]
    The collection of product graphs $\prodGraph=(\verts_\prodGraph,\edges_\prodGraph)$ contains all $i=1\dots\numTriangles{\shapeX}$ sets of product vertices $\verts_{\prodGraph_i}$ and sets of product edges $\edges_{\prodGraph_i}$ such that the vertices $\verts_\prodGraph$ and edges $\edges_\prodGraph$ of $\prodGraph$ read
    \begin{equation}
        \begin{aligned}
            \verts_\prodGraph &= \verts_{\prodGraph_1}\cup \dots \cup \verts_{\prodGraph_\numTriangles{\shapeX}},\\
            \edges_\prodGraph &= \edges_{\prodGraph_1}\cup \dots \cup \edges_{\prodGraph_\numTriangles{\shapeX}}.
        \end{aligned}
    \end{equation}
\end{definition}

By definition of the surface cycles and the product graphs, the product edges $\edges_\prodGraph$ of the product graph collection $\prodGraph$ encode all potential matchings between edges of shape $\shapeX$ and edges/vertices of shape $\shapeY$.

As mentioned earlier, our goal is to derive an integer linear programming formalism for the $\numTriangles{\shapeX}$-many shortest cyclic path problems collected in $\prodGraph$ such that we can incorporate additional constraints.
To this end, we encode the $k$-th product edge $e_k\in\edges_\prodGraph$ with a binary variable $x_k\in\{0,1\}$ where $x_k=1$ means that product edge $e_k$ is part of the computed matching.
Furthermore, and to encode that matched edges must form a connected path in $\prodGraph$, we consider continuity constraints which read
\begin{equation}\tag{CONT}\label{eq:continuity}
    \forall\; v \in \verts_\prodGraph:  
            \sum_{k: e_k = (\bullet, v)\in \edges_\prodGraph} x_k 
            = 
            \sum_{j: e_j = (v, \bullet)\in \edges_\prodGraph} x_j.
\end{equation}
Intuitively, whenever a product vertex $v\in\verts_\prodGraph$ has an `active' incoming edge, it also needs to have an `active' outgoing edge.
Consequently, continuity and with that geometric consistency along the resulting path is preserved.
Yet, geometric consistency between the independent subproblems is not yet ensured, i.e.~the matchings of neighbouring triangles of shape $\shapeX$ are not necessarily neighbouring on shape $\shapeY$.

To account for that, we adapt the idea presented in~\cite{roetzer2025geco} and couple the independent subproblems at opposite product edges whenever a product edge amounts to non-boundary edges on shape $\shapeX$ and shape $\shapeY$. 
To this end, we consider the following coupling constraints
\begin{equation}\tag{COUPL}\label{eq:coupling}
    \forall\; kj; \; e_k e_j \in \edges_{\prodGraph}^\interior; \; e_k=-e_j:\quad x_k = x_j.
\end{equation}
Here, $\edges_{\prodGraph}^\interior \subset \edges_\prodGraph$ are all product edges of $\prodGraph$ which solely consist of non-boundary vertices on shape $\shapeX$ and a non-boundary vertices on shape $\shapeY$, i.e.~$\edges_{\prodGraph}^\interior$ reads
\begin{equation}
    \begin{aligned}
        \edges_{\prodGraph}^\interior = \{e=\left(\twovector{x}{y}, \twovector{\bar{x}}{\bar{y}}\right)\in\edges_\prodGraph \; | \; 
        &x,\bar{x}\in\verts_\shapeX^\interior,\\
        &y,\bar{y}\in\verts_\shapeY^\interior\; \}.
    \end{aligned}
\end{equation}
These coupling constraints ensure that the neighbourhood of every surface cycle (which does not contain a boundary vertex) is preserved.
We note that it is important to enforce the coupling only on non-boundary product edges. This allows us to handle the fact that the overlap is unknown, i.e. it would not be possible to match only parts of the shape if the coupling were also defined for boundary edges.

While the above constraints \eqref{eq:continuity} and \eqref{eq:coupling} effectively enforce geometric consistency, we also want to ensure that resulting matchings are injective (i.e.~every vertex of shape $\shapeX$ within the overlapping region should be matched) and surjective (i.e.~every vertex of shape $\shapeY$ within the overlapping region should be matched), which we discuss next.

\subsection{Injectivity and Surjectivity of Overlap}\label{sec:inj-surj}

Our goal is to have a matching which is injective and surjective within the overlapping region between both shapes, i.e.~we want to match every vertex of $\shapeX$ and every vertex of $\shapeY$ whenever these vertices are in the overlapping region.
Yet, we do not know the overlapping region a priori.
To account for that, we introduce variables $\injectivitySlack \in \{0,1\}^{|\edges_\shapeX|}$ and $\surjectivitySlack \in \{0,1\}^{|\verts_\shapeY|}$.
Using $\injectivitySlack$, our injectivity constraints read
\begin{equation}\tag{INJY}\label{eq:injectivity}
    \forall\;j: e_j= (x, \bar{x}) \in \edges_{\shapeX}: 
    \!\!
    \sum_{k: e_k=\left(\twovector{x}{\bullet},\twovector{\bar{x}}{\bullet}\right)\in\edges_{{\prodGraph}}}
    \!\!
    x_k - \injectivitySlack_j = 1.
\end{equation}
Hence, we enforce injectivity by requiring that every edge $e_j\in\edges_\shapeX$ of shape $\shapeX$ is matched exactly once whenever $e_j$ is in the overlapping region (i.e.~whenever $\injectivitySlack_j=0$).
Furthermore, our surjectivity constraints read
\begin{equation}\tag{SURJY}\label{eq:surjectivity}
    \begin{aligned}
        \forall\;j:\; v_j \in &\verts_{\shapeY},\; v_j=y \;\lor\; v_j =\bar{y}: \\
        &\sum_{k: e_k=\left(\twovector{\bullet}{y},\twovector{\bullet}{\bar{y}}\right)\in\edges_{{\prodGraph}}}
        \!\!
        x_k - \surjectivitySlack_j \geq 1.
    \end{aligned}
\end{equation}
Similarly to injectivity, we enforce surjectivity by requiring that every vertex $v_j\in\verts_\shapeY$ is matched at least once whenever $v_j$ is in the overlapping region (i.e.~whenever $\surjectivitySlack_j=0$).

With all of the required constraints now at hand, we can define our integer linear program (ILP) for geometrically consistent partial-partial shape matching.

\subsection{ILP for Partial-Partial Shape Matching}\label{sec:resulting-ilp}
In the following, we state our ILP for geometrically consistent partial-partial shape matching. 
To this end, we collect all binary variables $x_k$ (which encode if product edge $e_k\in\edges_\prodGraph$ is part of the final matching) in a binary vector $x\in\{0,1\}^{|\edges_\prodGraph|}$.
Furthermore, we define a matching cost $\matchingCost_k$ for every product edge $e_k\in\edges_\prodGraph$ and also collect all matching costs in a vector $\matchingCost\in\mathbb{R}^+$.
For the matching cost, we compute feature-differences of per-vertex features, see \cref{sec:experiment-setup}.
Additionally, we compute probabilities if an edge, respectively vertex, of shape $\shapeX$, respectively shape $\shapeY$, is within the overlapping region and collect these in vectors $\overlapX\in[0,1]^{|\edges_\shapeX|}$, respectively $\overlapY\in[0,1]^{|\verts_\shapeY|}$, see \cref{sec:experiment-setup}. 

With $\matchingCost, \overlapX$, and $\overlapY$, we can state our ILP for geometrically consistent partial-partial shape matching.

\begin{equation}\tag{PP-ILP}\label{eq:partial-partial-ILP}
    \begin{aligned}
        &\underset{x, \injectivitySlack,  \surjectivitySlack}{\min} 
        \matchingCost^T x + \lambda\overlapX^T\injectivitySlack + \lambda\overlapY^T\surjectivitySlack\\
        &\;\text{s.t. } \eqref{eq:continuity}, \eqref{eq:coupling}, \eqref{eq:injectivity}, \eqref{eq:surjectivity},\\
        &\;\;\quad x \in\{0,1\}^{|\edges_\prodGraph|}, \injectivitySlack\in\{0,1\}^{|\edges_\shapeX|},  \surjectivitySlack \in\{0,1\}^{|\verts_\shapeY|}
    \end{aligned}
\end{equation}
Here, $\lambda$ is a weighting term between matching costs and overlap probabilities,
see \cref{sec:experiment-setup}.

Even though \eqref{eq:partial-partial-ILP} is an integer linear formalism for partial-partial shape matching, it involves many binary variables and thus is hard to solve.
This might limit its scalability (while still scaling significantly better than previous methods, cf. \cref{fig:teaser}). 
Thus, to match higher resolution shapes, we solve \eqref{eq:partial-partial-ILP} on lower shape resolutions and propagate solutions to higher shape resolutions, which we discuss next.

\subsection{Scaling to Higher Resolutions}\label{sec:coarse-to-fine}

\begin{figure}
    \centering
    \includegraphics[width=0.9\columnwidth]{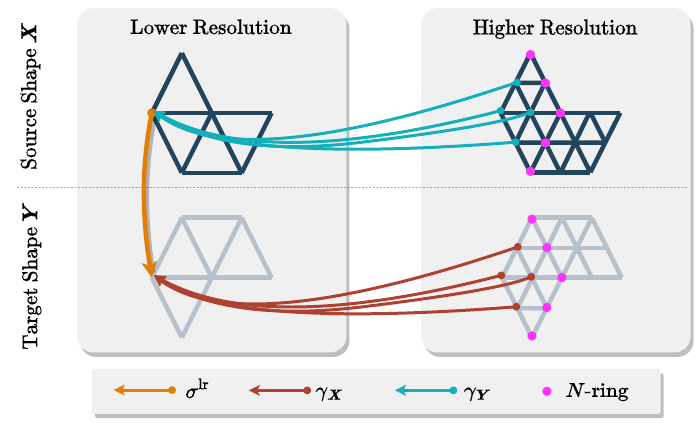}
    \caption{We \textbf{prune} instances of \eqref{eq:partial-partial-ILP} for pairs of higher resolution shapes by using computed matching of (the same pair of) lower resolution shapes. In particular, on higher resolution shapes, we only allow for pairs of vertices which map to a matching on lower resolutions shapes within their $N$-ring neighbourhood (we show $N=1$ in the figure), i.e.~in this example, all coloured vertices on higher resolution shape $\shapeX$ can be matched to all coloured vertices higher resolution shape $\shapeY$.}
    \label{fig:pruning}
\end{figure}

Due to many binary variables and complex constraints, \eqref{eq:partial-partial-ILP} might only be trackable to solve for lower resolution shapes. 
To scale to higher resolution shapes, we use a solution found on lower resolution shapes to prune \eqref{eq:partial-partial-ILP} for higher resolution shapes~\cite{ehm2024geometrically, windheuser2011large}.
In particular, by solving \eqref{eq:partial-partial-ILP}, we compute a mapping $\sigma^\lowres : \verts_\shapeX^\lowres \rightarrow \verts_\shapeY^\lowres$ on lower resolution variants of shapes $\shapeX$ and $\shapeY$, i.e.~$\verts_\shapeX^\lowres \subset \verts_\shapeX$ and $\verts_\shapeY^\lowres \subset \verts_\shapeY$.
Additionally, we maintain a mapping from higher resolution vertices to lower resolution vertices, i.e.~$\highToLowresX : \verts_\shapeX \rightarrow \verts_\shapeX^\lowres$ and $\highToLowresY : \verts_\shapeY \rightarrow \verts_\shapeY^\lowres$.
Furthermore, we consider $N\text{-ring}(\cdot)$ to be the set of vertices which lie within the $N$-ring neighbourhood of a vertex on a shape (i.e.~all vertices that are at most $N$ edges away from a given vertex).
With that, we can compute a set $\AllowedMatches$ consisting of product vertices which comprise the allowed matching pairs on higher resolution shapes.
\begin{equation}
    \begin{aligned}
        \AllowedMatches \coloneqq \bigl\{\twovector{x}{y} \;|\;&
        x \in \verts_\shapeX,\; y \in \verts_\shapeY,
        \\
        &\gamma_\shapeY(N\text{-ring}(y)) \cap \sigma^\lowres(\gamma_\shapeX(N\text{-ring}(x)) \neq \emptyset \bigr\}
    \end{aligned}
    \label{eq:upsample}
\end{equation}
In a nutshell, $\AllowedMatches$ consists of all pairs of vertices (including their  $N$-ring neighbourhood) of higher resolution shapes which map to a pair of matched vertices on lower resolution shapes, see \cref{fig:pruning} for an illustration.

For higher resolution shapes, we employ the following additional constraints in \eqref{eq:partial-partial-ILP} to prune matchings which do not lie within the $N$-ring neighbourhood of the lower resolution matching $\sigma^\lowres$.
\begin{equation}\tag{PRUNE}
    \forall\; k: e_k=\left(v,\bar{v}\right) \in\edges_\prodGraph,\; v\notin \AllowedMatches\land \bar{v}\notin \AllowedMatches: x_k = 0
\end{equation}

We note that the above described pruning and subsequent solving of \eqref{eq:partial-partial-ILP} can be repeated multiple times, such that we can go from lower resolution to higher resolution shapes via intermediate resolutions.

\section{Experiments}
\label{sec:experiments}

In this section, we empirically evaluate our method's performance. 
To this end, we first introduce the datasets and comparison methods as well as the experimental setup that we use for our method and competitors. 
After that, we explain the metrics that we use.
Finally, we report quantitative and qualitative matching results. 

\subsection{Dataset \& Competing Methods \& Setup}\label{sec:experiment-setup}

\textbf{Datasets.}
We conduct a quantitative comparison on two datasets.
First, we use CP2P24~\cite{ehm2024partial}, which builds on SHREC16 CUTS~\cite{cosmo2016shrec} and defines a new test set for CP2P~\cite{attaiki2021dpfm} such that the test set does not contain shapes which exist in the SHREC16 CUTS train set (we report results only on test set shapes). 
The data set consists of 76 isometric deformed animal and human shapes with varying partiality ranging from $10\%$-$90\%$ overlap.
Second, we use PSMAL~\cite{ehm2024partial}, which consists of 49 remeshed animal shapes from 8 different species from SMAL~\cite{zuffi20173d} and where partiality is introduced by cutting shapes in half using planes in 3D space, see~\cite{ehm2024partial} for more details.
Similar to~\cite{cao2023unsupervised}, we report results only on the test set of SMAL~\cite{zuffi20173d}.

\textbf{Competing Methods.} 
We compare our method against three other approaches. 
DPFM~\cite{attaiki2021dpfm} is a supervised learning-based approach that uses a cross-attention feature refinement network for partial-to-partial shape matching, predicting both the overlapping region and features for every vertex.
EchoMatch~\cite{xie2025echomatch} is a learning-based method which uses features computed from image foundation models~\cite{dutt2024} as input, feeds them into a DiffusionNet~\cite{sharp2022diffusionnet}, and predicts overlapping regions based on so-called correspondences reflections.
GC-PPSM~\cite{ehm2024partial} builds its formalism on the integer linear program for geometrically consistent 3D shape matching introduced in~\cite{windheuser2011geometrically} but is specifically tailored for geometrically consistent partial-partial 3D shape matching.
As such, GC-PPSM uses a non-linear integer programming formalism.

\textbf{Setup.} 
For all axiomatic methods, we use features of EchoMatch~\cite{xie2025echomatch} to compute matching costs, i.e.~we compute the cosine distance between the features of respective vertices of the product edges.
Additionally, we use the overlap predictions $\overlapX$ and $\overlapY$ of EchoMatch in our method.
For our approach, we solve lower resolution pairs of shapes on a combined number of $600$ triangles and use these solutions to prune higher resolution pairs with a combined resolution of $1000$ triangles, for which we use the $N=2$ ring neighbourhood.
To this end, we downsample shapes using algorithms provided in~\cite{libigl} such that shapes have similarly sized triangles.
We solve \eqref{eq:partial-partial-ILP} using off-the-shelf solver Gurobi~\cite{gurobi} where we set a time limit of $60$ for resolutions $600$ and then upsample to $1000$ triangles via $800$ triangles, where every upsample step has a time limit of $30$ minutes.
We set $\lambda=0.5$ for PSMAL and $\lambda=0.3$ for CP2P24.
Furthermore, for GC-PPSM~\cite{ehm2024partial}
we use the experimental setup as depicted in the paper.
To make geodesic errors comparable, we upsample solutions to full resolutions using the upsampling technique proposed in \cite{ehm2024partial} with a neighbourhood size of $1$ and choose the direction of the matching with the better objective value of the \eqref{eq:partial-partial-ILP}.

\subsection{Metrics}

\textbf{Intersection over Union.}
For each vertex on one shape, e.g.~for $\shapeX$, we can define a matching vector $m\in \{0,1\}^{(|\verts_\shapeX|)}$ (by evaluating the computed matching of our method) as well as obtain a ground truth vector $g \in \{0,1\}^{(|\verts_\shapeX|)}$ (by evaluating the ground-truth correspondences).
Both vectors $m$ and $g$ indicate whether a vertex $x_i\in\verts_\shapeX$ lies within the overlapping region, i.e.~$g_i=1$ respectively $m_i=1$, or does not lie within the overlapping region, i.e.~$g_i=0$ respectively $m_i=0$.
Using these two vectors $m, g$, we follow~\cite{ehm2024partial} and use the intersection over Union (IoU), i.e.~$\text{IoU} = \frac{|m \cap g|}{|m \cup g|}$, to quantify the accuracy of overlap region predictions.
We report the mean IoU (mIoU) over all shapes.

\textbf{Geodesic Error.}
We measure correspondence quality on the intersection of predicted and overlapping region using the geodesic error, for which we follow the Princeton Protocol~\cite{kim2011blended} and normalise distances by the shape diameter of the full shape.

\textbf{Dirichlet Energy}
To evaluate matching smoothness, we follow~\cite{roetzer2024spidermatch} and report Dirichlet Energies. 
To this end, we rigidly align shapes using the ground-truth matching, compute a deformation field between matched points, and evaluate the smoothness of this deformation field according to the Laplacian of the target shapes, see~\cite{roetzer2024spidermatch} for more details.

\textbf{Geodesic Edge Distortion (\shortNewMetric)}
Dirichlet Energies compute an extrinsic quantity and, with that, depend on the amount of non-rigid transformation between both shapes (even for isometric deformed shapes). 
To circumvent this issue, we propose a new metric that better captures the intrinsic neighbourhood changes between two shapes. 
In particular, for every edge $e=(x, \bar{x})\in\edges_\shapeX$ on shape $\shapeX$, we compute the geodesic distance $d_\shapeY(\sigma(x),\sigma(\bar{x}))$ on $\shapeY$ where $\sigma : \verts_\shapeX \rightarrow \verts_\shapeY$ is the computed matching.

\subsection{Results}

\textbf{Runtime.}
In \cref{fig:teaser} right, we evaluate the runtime of our method (without pruning) over 10 instances of the CP2P24 dataset.
We compare to the only other geometrically consistent shape matching, which is specifically designed for partial-partial shape matching.
We can see that our method scales better compared to~\cite{ehm2024partial}, very likely due to our integer linear programming formalism (opposed to the non-linear integer programming formalism proposed in~\cite{ehm2024partial}).

\begin{table}[htp]
    \centering
    \small
    \setlength{\tabcolsep}{0.25em}
    \begin{tabular}{c}
            \textsc{CP2P24}\\
        \begin{tabular}{l|c c c c}
            \toprule
            Method & IoU ($\uparrow$) & GeoError ($\downarrow$) & Dirichlet ($\downarrow$) & \shortNewMetric~($\downarrow$)\\
            \midrule
            EchoMatch & \underline{84.72} & \underline{2.29} & \underline{57.83} & 6.15\\
            DPFM &74.17 & 3.04 & 65.03 & 4.08\\
            GC-PPSM & 68.15 & 8.87 & 93.20 & \textbf{1.58} \\
            \midrule
            \textbf{Ours} & \textbf{85.28} & \textbf{2.23} & \textbf{44.94} & \underline{1.69}\\
            \bottomrule
    
        \end{tabular}\\
        \vspace{-0.2cm}\\
        \textsc{PSMAL}\\
        \begin{tabular}{l|c c c c}
            \toprule
            Method & IoU ($\uparrow$) & GeoError ($\downarrow$) & Dirichlet ($\downarrow$) & \shortNewMetric~($\downarrow$)\\
            \midrule
            EchoMatch & \underline{84.72} %
            & \underline{4.01} & \underline{88.05} & 7.75\\
            DPFM & 73.63 %
            & 9.72 & 130.44 & 4.13\\
            GC-PPSM &62.27 & 11.74 & 93.12 & \textbf{1.87}\\
            \midrule
            \textbf{Ours} & \textbf{84.96} & \textbf{3.87} & \textbf{34.52} & \underline{2.01}\\
            \bottomrule
        \end{tabular}\\
    \end{tabular}
    \caption{\textbf{Comparison of quantitative results ($\times 100$) on the CP2P24 and PSMAL dataset} in terms of smoothness, overlap prediction and correspondence quality.
    We indicate \textbf{best} and \underline{second best} results.
    }
    \label{tab:cp2p_quantitative}
\end{table}

\textbf{Quantitative Evaluation.} 
On both datasets PSMAL and CP2P24, we observe that learning-based methods produce less smooth results according to Dirichlet energies as well as \shortNewMetric, see \cref{tab:cp2p_quantitative}.
Additionally, our approach outperforms the other learning-based methods in predicting overlapping regions (mIoU) and in correspondence prediction quality (GeoErrors).

Although GC-PPSM achieves better performance in terms of \shortNewMetric, we emphasise that such metrics must always be interpreted in conjunction with overlap prediction, as all other metrics are evaluated on the intersection of the ground truth and the predicted overlapping region.
\newcommand{\horseSource}{\includegraphics[width=0.07\linewidth]{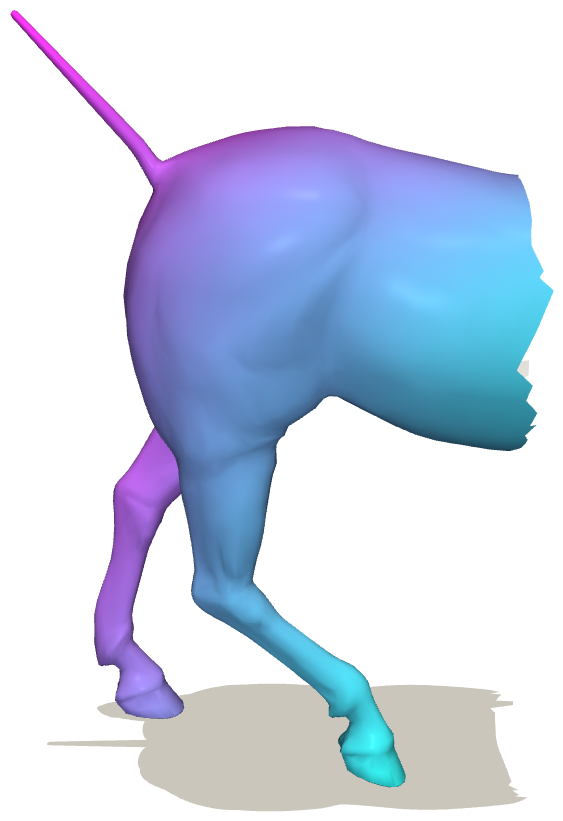}}
\newcommand{\horseResult}[1]{\includegraphics[width=0.14\linewidth]{vis/horse-6_horse-13/#1.png}}
\newcommand{\dogSource}{\includegraphics[width=0.1\linewidth]{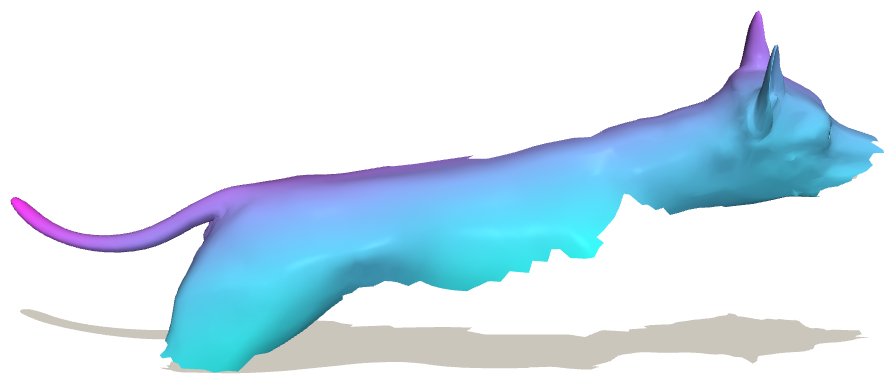}}
\newcommand{\dogResult}[1]{\includegraphics[width=0.05\linewidth]{vis/dog-15_dog-6/#1.png}}
\newcommand{\psmalFirstSource}{\includegraphics[width=0.08\linewidth]{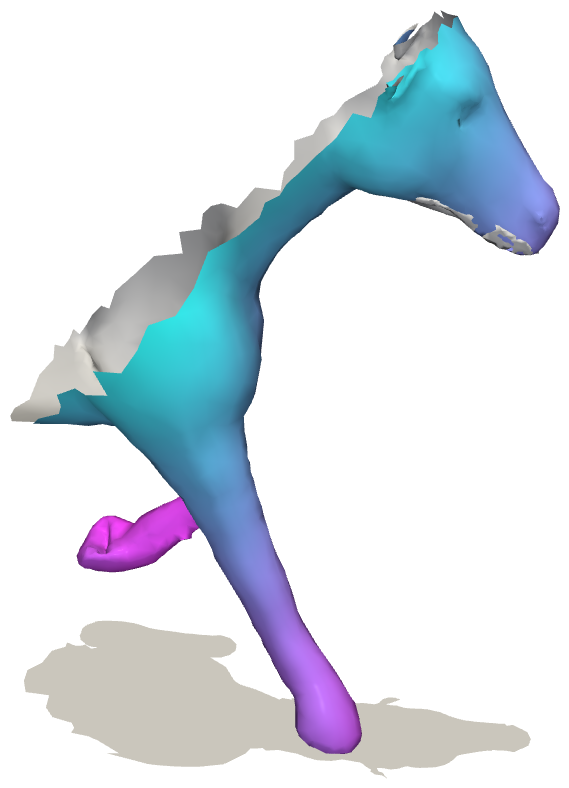}}
\newcommand{\psmalFirstOther}[1]{\includegraphics[width=0.12\linewidth]{vis/cuts_11_horse_07_cuts_3_hippo_03/#1.png}}
\newcommand{\psmalSecondSource}{\includegraphics[width=0.07\linewidth]{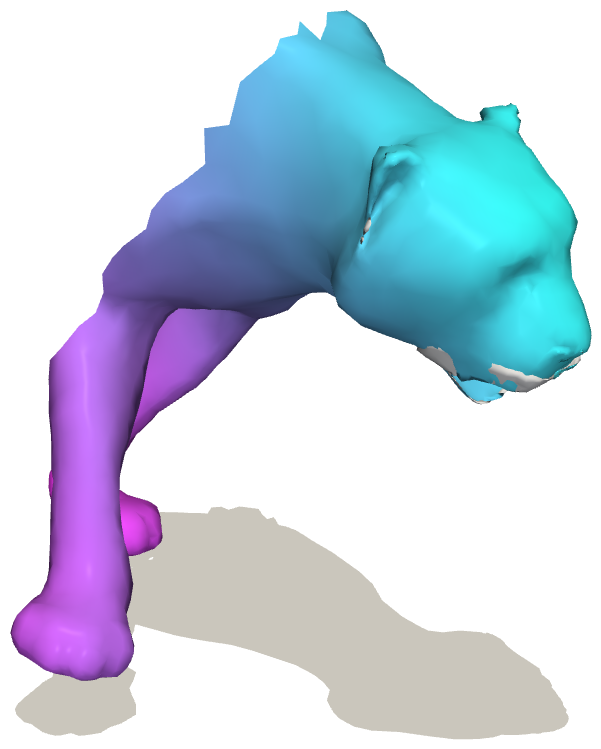}}
\newcommand{\psmalSecondOther}[1]{\includegraphics[width=0.115\linewidth]{vis/cuts_22_cougar_02_cuts_4_horse_01/#1.png}}
\begin{figure*}[h]
     \centering
     \small
     \renewcommand{\arraystretch}{0.01}
      \setlength{\tabcolsep}{0.7em}
     \begin{tabular}{ccccccc}%
     & \small{Source} & \small{GT} & \small{GC-PPSM}~\cite{ehm2024partial} & \small{DPFM}~\cite{attaiki2021dpfm} & \small{Echo}~\cite{xie2025echomatch} & \small{Ours}\\
     \toprule
     \multirow{ 2}{*}{
        \rotatebox{90}{\small{CP2P24}\hspace{-0.5cm}}
     }
     & \horseSource & \horseResult{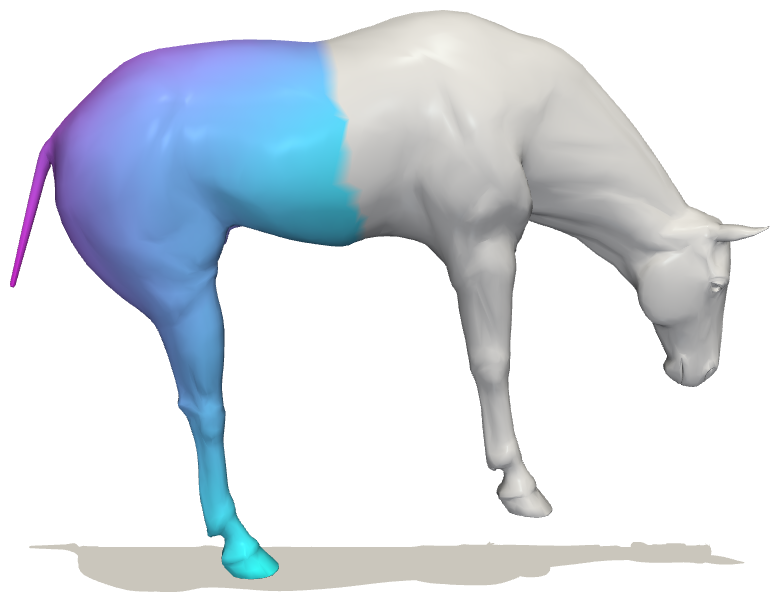} & \horseResult{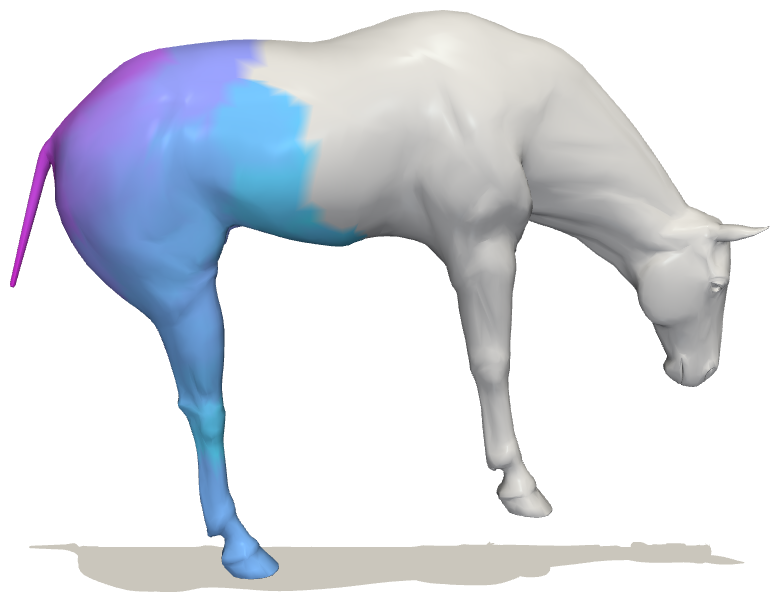} & \horseResult{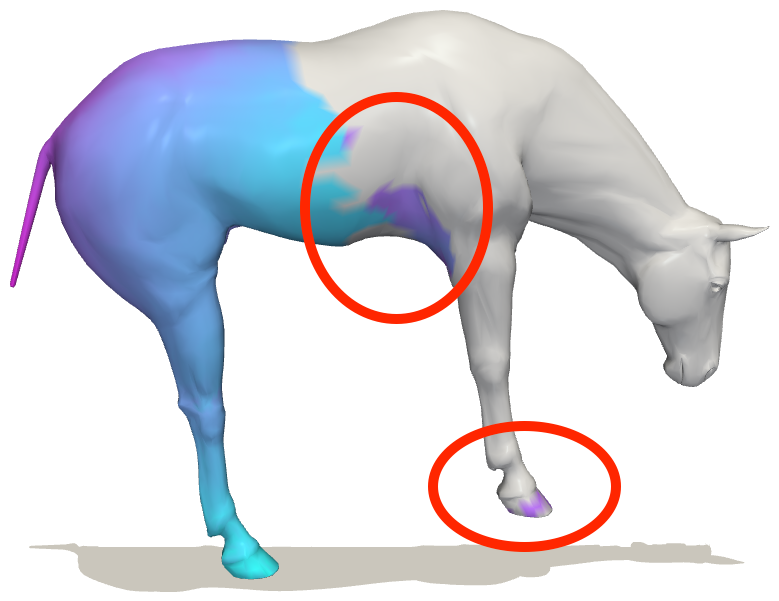}  &\horseResult{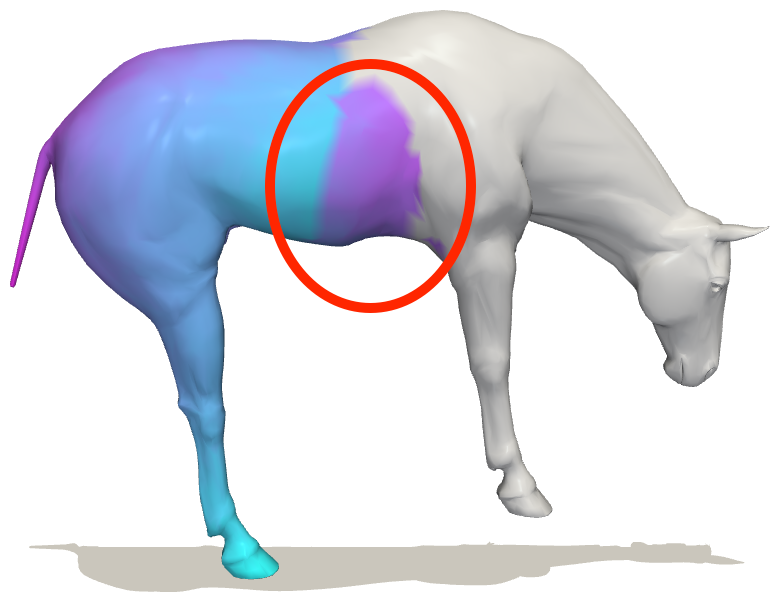} &\horseResult{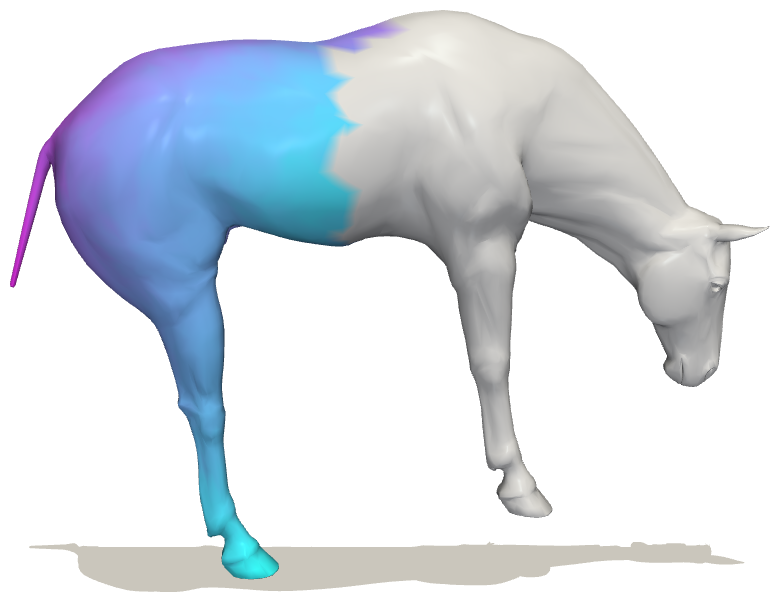} \\
    & \dogSource & \dogResult{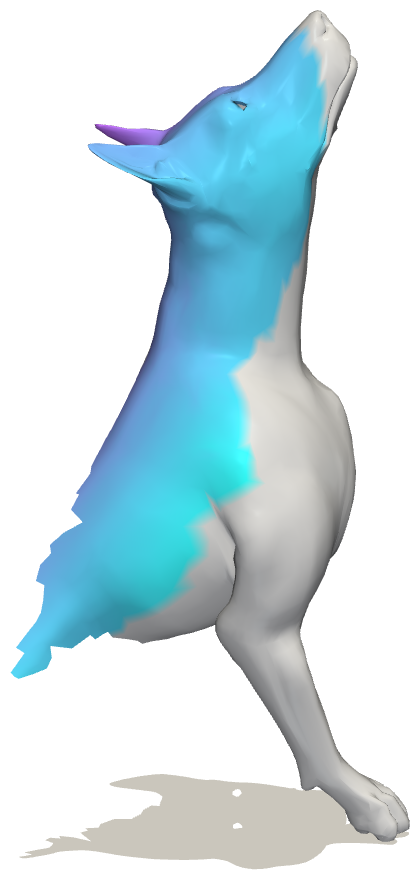} & \dogResult{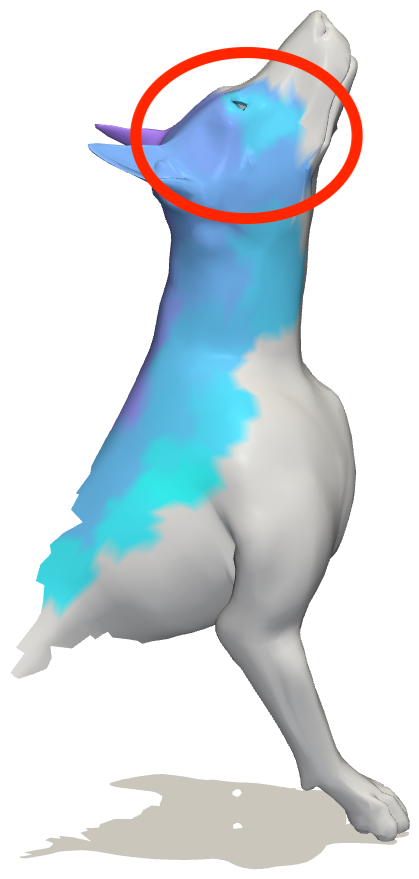} & \dogResult{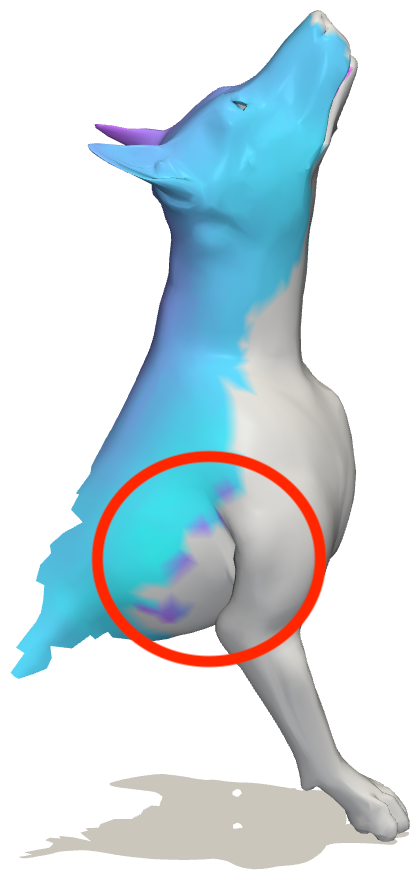}  &\dogResult{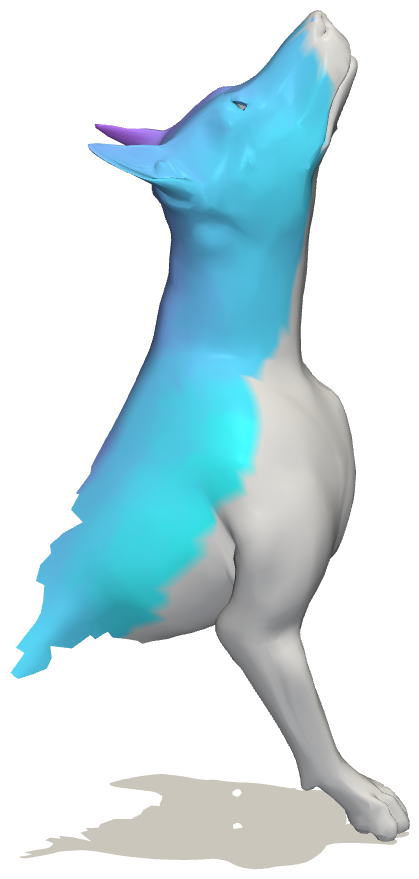} &\dogResult{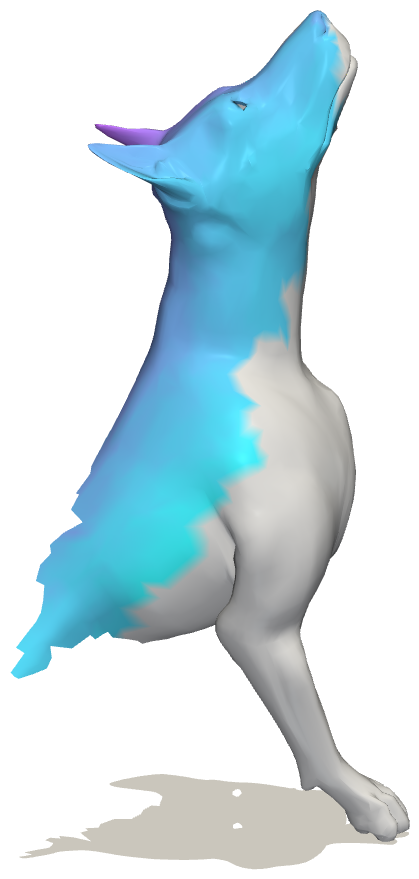} \\
    \midrule
     \multirow{ 2}{*}{
        \rotatebox{90}{\small{PSMAL}\hspace{-0.5cm}}
     }
     & \psmalFirstSource & \psmalFirstOther{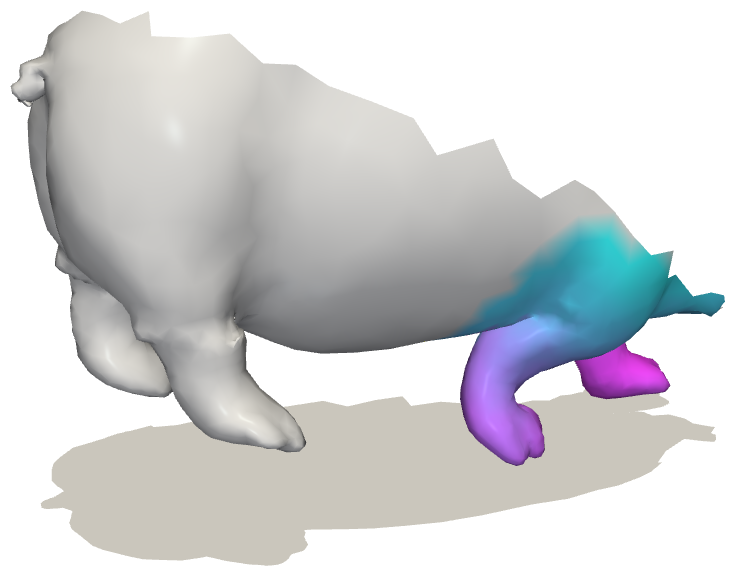} & \psmalFirstOther{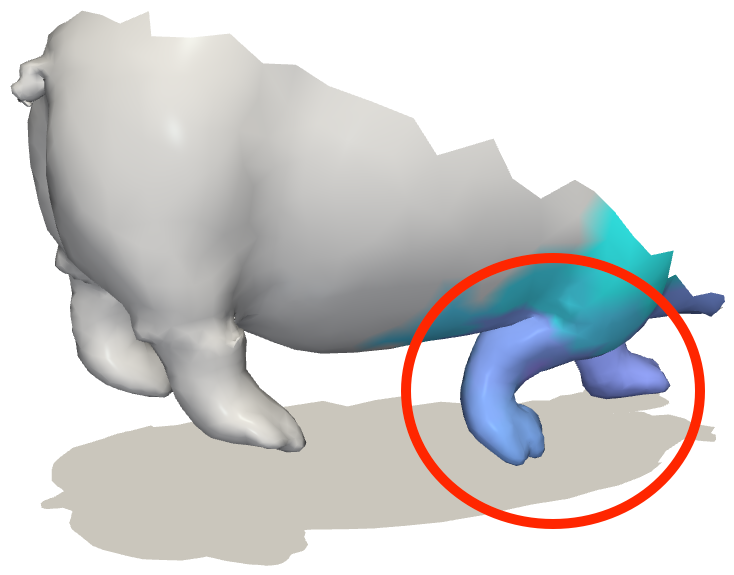} & \psmalFirstOther{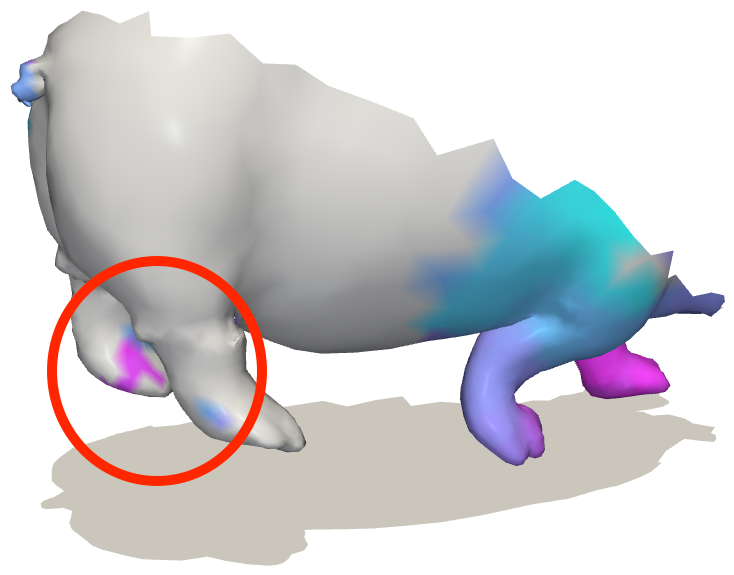}  &\psmalFirstOther{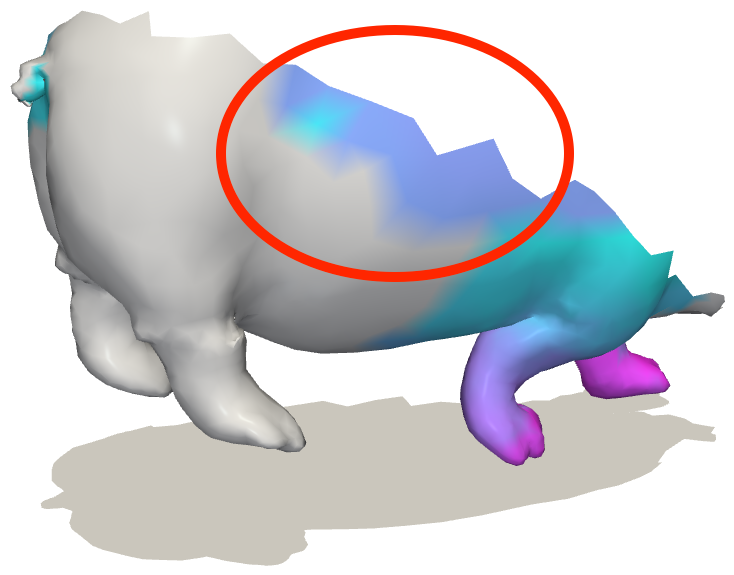} &\psmalFirstOther{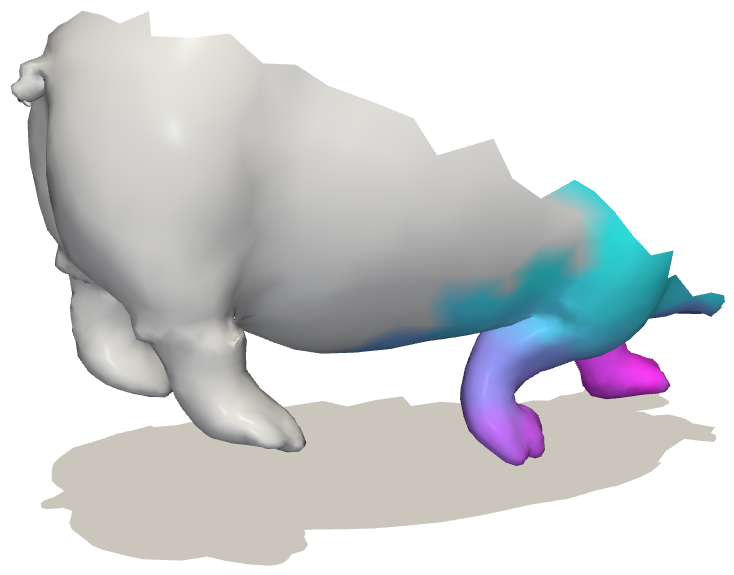} \\
     & \psmalSecondSource & \psmalSecondOther{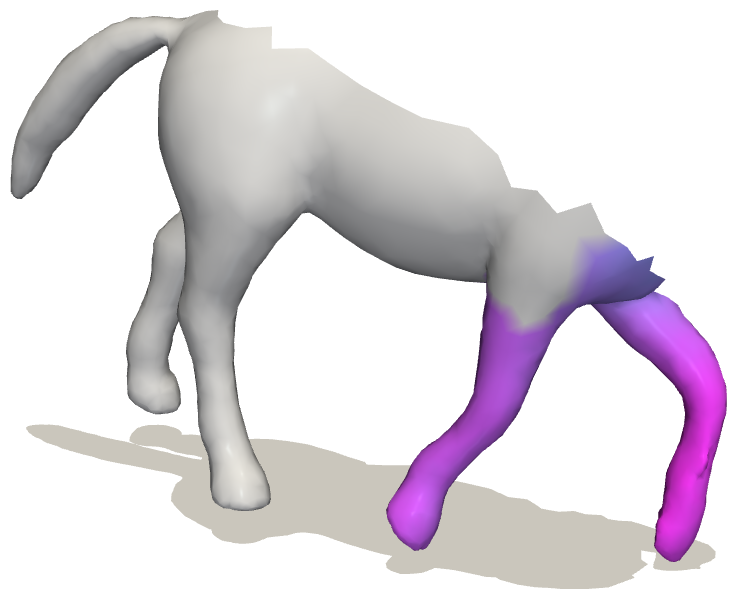}  &\psmalSecondOther{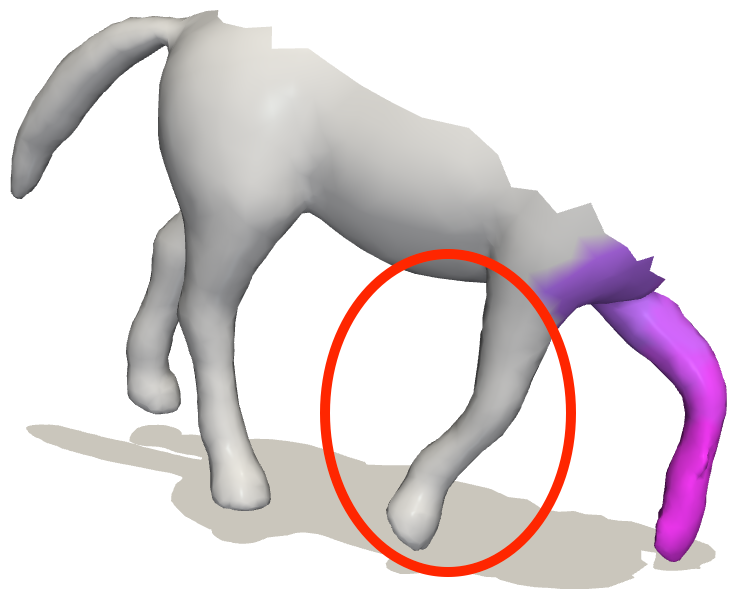} & \psmalSecondOther{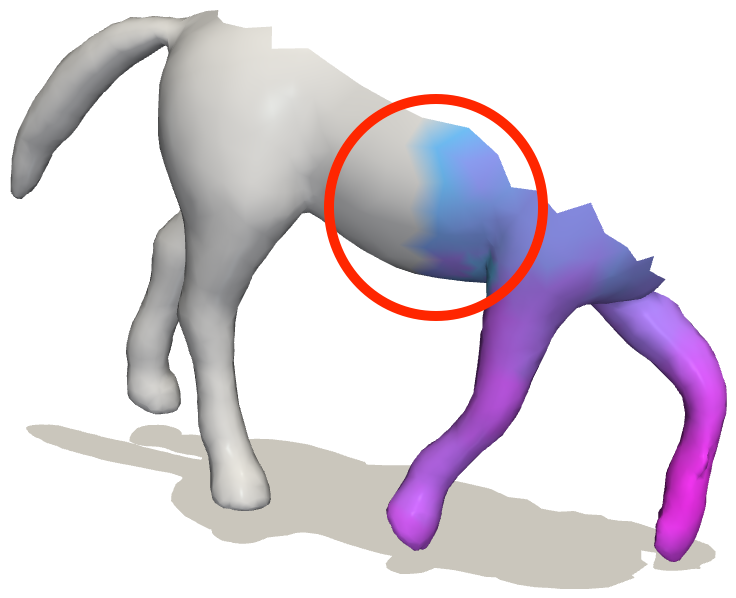}  &\psmalSecondOther{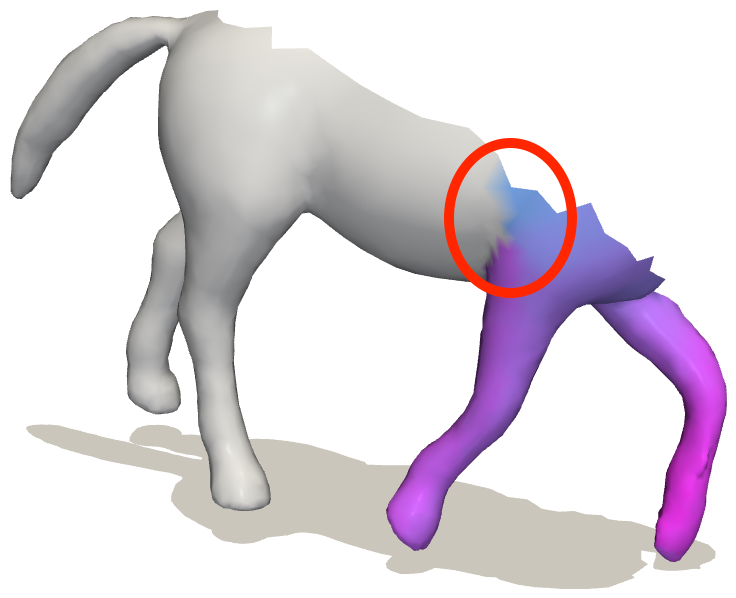} &\psmalSecondOther{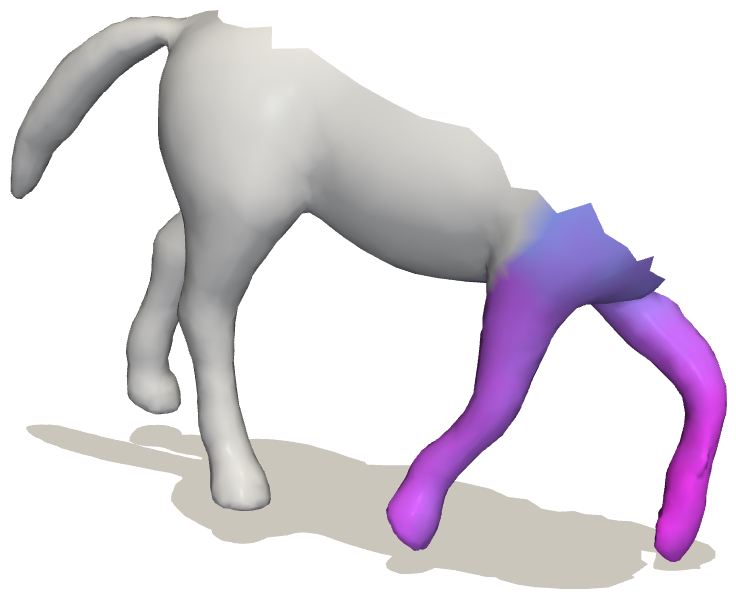} \\
     \bottomrule
     \end{tabular}
     \caption{%
     We show \textbf{qualitative results} and the ground-truth (GT) of shape pairs of CP2P24 (first two rows) and PSMAL (last two rows) using methods GC-PPSM~\cite{ehm2024partial}, DPFM~\cite{attaiki2021dpfm}, Echo~\cite{xie2025echomatch} and ours.
     We can see that our method leads to overall smoother matching results.
    }
     \label{fig:qualitative_results}
 \end{figure*}

\textbf{Qualitative Evaluation.}
We present qualitative results in \cref{fig:qualitative_results}.
DPFM mainly struggles with overlap prediction (see rows 3 and 4), while EchoMatch produces inaccurate correspondence predictions (row 3).
GC-PPSM exhibits patch-wise inconsistencies, such as in row 2, which can be attributed to solving at a very low resolution.
In contrast, our method generally yields smooth and accurate matchings, with minor inaccuracies near boundaries, likely due to the discretisation artefacts from the initial low-resolution.

\subsection{Ablation Studies}
We use 10 random samples of the train set of the CP2P~\cite{attaiki2021dpfm} and the PSMAL~\cite{ehm2024partial} to determine the best parameter for the respective dataset.
We show ablation studies for the neighbourhood size when upsampling and the weighting factor in the overlapping region in \cref{tab:ablation-lambda} in the supplementary.

\section{Discussion \& Limitations}
\label{sec:discussion}

Our approach is the first integer linear programming formalism for partial-partial shape matching, which only allows for geometrically consistent solutions.
Nevertheless, since we build our formalism on GeCo~\cite{roetzer2025geco}, the solution space contains undesirable solutions, such as e.g.~orientation flips.
Furthermore, resulting matching results might contain multiple, disconnected overlapping regions.
Yet, our method empirically shows good performance in terms of matching errors, smoothness and overlap scores. 

Apart from that, even though our approach is an integer linear programming formalism (and not a non-linear integer programming formalism compared to previous work~\cite{ehm2024partial}), it remains computationally demanding in practice.
Consequently, worst-case runtimes remain exponential and limit the scalability of our method.
Nevertheless, our approach shows better scalability compared to previous works and we consider our formalism an important step towards making partial-partial shape matching more accessible.

\section{Conclusion}
\label{sec:conclusion}

We have presented the first integer \emph{linear} programming formalism explicitly tailored to the challenging task of partial-partial 3D shape matching, i.e.~our formalism yields neighbourhood preserving matchings of an unknown overlapping region between two partial shapes.
Furthermore, by using geometric consistency as a strong prior, our approach allows to curate overlap predictions from other, data-driven approaches as e.g.~learning-based methods.
In addition, our formalism is more scalable than previous, non-linear integer programming approaches and shows favourable results in terms of matching quality, smoothness, and overlap scores.

Overall, there exists only few methods tackling partial-partial 3D shape matching 
and there are still several unsolved challenges.
In that sense, we consider our work to be an important contribution for the shape matching community
and we hope to inspire follow-up works dealing with partial 3D shapes and partial data in general.

\section{Acknowledgements}
This work was supported by the ERC Advanced Grant “SIMULACRON” (agreement \#884679) and the ERC Starting Grant “Harmony” (agreement \#101160648).

{
    \small
    \bibliographystyle{ieeenat_fullname}
    \bibliography{main}
}
\clearpage
\setcounter{page}{1}
\maketitlesupplementary

\section{Ablation Studies}
In the following, we show ablation studies on the weighting factor $\lambda$, which weighs the matching cost against the overlap probabilities in \eqref{eq:partial-partial-ILP}, and on the neighbourhood size $N$, which influences the size of the allowed matching set, see \cref{eq:upsample}.

\subsection{Weighting of Overlap Prediction}
We analyse the mean IoU for different weighting factors $\lambda$ between $0$ and $1$ for $10$ random shape pairs for datasets CP2P, respectively PSMAL.
In Table~\ref{tab:ablation-lambda}, we show that $\lambda=0.3$, respectively $\lambda=0.5$, yields the best results on CP2P, respectively PSMAL.

\begin{table}[ht]
    \centering
    \small
    \begin{tabular}{c|ccccc}
    \toprule
    $\lambda$ & 0 & 0.1 & 0.3 & 0.5 & 1.0 \\
    \midrule
    CP2P24 & 0 & 72.01 & \textbf{87.40} & 86.53 & 84.65  \\
    PSMAL & 0 & 53.17 & 82.03 & \textbf{82.35} & 80.02\\
    \bottomrule
    \end{tabular}
    \caption{Ablation study of the mean IoU on \textbf{different weighting factors for the overlapping region} on $10$ examples of the CP2P/PSMAL train dataset on a combined number of $600$ faces for both shapes.}
    \label{tab:ablation-lambda}
\end{table}

\subsection{Neighbourhood Ring Size}
We compare different neighbourhood ring sizes $N$ on 10 random samples of the CP2P train set in terms of optimisation time and mean IoU in Table~\ref{tab:neighbourhood_size} for an upsampling step from $600$ to $800$ total triangles. We observe the best results with $N=2$.
\begin{table}[ht]
    \centering
    \small
    \begin{tabular}{c|ccccc}
    \toprule
    $N$ & 0 & 2 & 4 & 6 \\
    \midrule
    mIoU ($\uparrow$) & 83.65 & \textbf{88.45} & \textbf{88.45} & 87.05\\
    Opt. Time (seconds) ($\downarrow$) & \textbf{11.27} & 141.58 & 340.03 & 351.31\\
    \bottomrule
    \end{tabular}
\caption{Ablation study of mIoU and optimisation time in seconds on \textbf{upsampling neighbourhood size} $N$: We observe the best combination of optimisation time and mIoU for $N=2$.}
\label{tab:neighbourhood_size}
\end{table}

\end{document}